\documentclass[11pt]{article}

\usepackage[final]{acl}
\usepackage{times}
\usepackage{latexsym}
\usepackage{amsmath} 
\usepackage{csvsimple}  
\usepackage[table]{xcolor}

\usepackage{booktabs}
\usepackage{geometry}
\usepackage{amsthm}
\usepackage{stfloats}
\usepackage{caption}

\usepackage{xcolor}
\definecolor{msftBlack}{RGB}{0,0,0}

\usepackage[T1]{fontenc}

\usepackage[utf8]{inputenc}

\usepackage{microtype}

\usepackage{inconsolata}

\usepackage{graphicx}

\usepackage[shortlabels,inline]{enumitem}
\usepackage{tikz,lipsum}
\usepackage[most]{tcolorbox}
\usepackage{ragged2e}
\usepackage[dvipsnames]{xcolor}
\usepackage{amsmath}
\usepackage{booktabs}
\usepackage{tabularray}
\usepackage{arydshln}
\usepackage{stmaryrd}
\usepackage{marvosym}
\usepackage{colortbl}
\usepackage{multicol}
\usepackage{multirow}
\usepackage{float}
\usepackage{verbatim}
\usepackage{minted}
\usepackage{fancyvrb}

\usepackage{cleveref}
\crefname{section}{\S}{\S}
\crefname{table}{Table}{Tables}
\crefname{figure}{Fig.}{Figs.}
\crefname{algorithm}{Alg.}{}
\crefname{ALC@unique}{Line}{Lines}
\crefname{equation}{Eq.}{Eqs.}
\crefname{appendix}{App.}{Apps.}
\crefformat{section}{\S#2#1#3} 

\usepackage{soul}
\definecolor{tablegray}{RGB}{223, 242, 252}

\usepackage{todonotes}

\NewDocumentCommand{\prompt}{O{} +m}{%
\begin{tcolorbox}[
    coltitle=white,
    colframe=black,
    colback=black!5!white,
    boxrule=1pt,
    enhanced jigsaw,
    breakable,
    pad at break*=2mm,
    left=2pt,
    right=2pt,
    top=2pt,
    bottom=2pt,
    fontupper=\small,
    fontlower=\small,
    title={#1}, 
]
#2 
\end{tcolorbox}
}

\usepackage[tikz]{bclogo}
\newcommand{\finding}[1]{
  \begin{bclogo}[couleur=msftBlack!05, epBord=1, arrondi=0.2, logo=\bcetoile, marge=5, ombre=true, blur, couleurBord=msftBlack!10, tailleOndu=1, sousTitre={\em #1}]{}
  \end{bclogo}
}
\title{Read Between the Lines: A Benchmark for Uncovering Political Bias in Bangla News Articles}

\author{
Nusrat Jahan Lia \\
University of Dhaka \\
\texttt{bsse1306@iit.du.ac.bd} \\
\And
Shubhashis Roy Dipta \\
University of Maryland, \\
Baltimore County \\
\texttt{sroydip1@umbc.edu} \\
\And
Abdullah Khan Zehady \\
Cisco Systems \\
\texttt{azehady@cisco.com} \\
\AND
Naymul Islam \\
BanglaLLM \\
\texttt{naymul504@gmail.com} \\
\And
Madhusodan Chakraborty \\
Maharishi International University \\
\texttt{opuchakraborty@gmail.com} \\
\And
Abdullah Al Wasif \\
Unityflow AI \\
\texttt{wasif@unityflow.ai} \\
}
\usepackage{xspace}
\newcommand{\ours}{BanglaBias\xspace} 

\begin{document}
\maketitle
\begin{abstract}
Detecting media bias is crucial, specifically in the South Asian region. Despite this, annotated datasets and computational studies for Bangla political bias research remain scarce. Crucially because, political stance detection in Bangla news requires understanding of linguistic cues, cultural context, subtle biases, rhetorical strategies, code-switching, implicit sentiment, and socio-political background. To address this, we introduce \ours, the first benchmark dataset of 200 politically significant and highly debated Bangla news articles, labeled for government-leaning, government-critique, and neutral stances, alongside diagnostic analyses for evaluating large language models (LLMs). Our comprehensive evaluation of 28 proprietary and open-source LLMs shows strong performance in detecting government-critique content (F1 up to 0.83) but substantial difficulty with neutral articles (F1 as low as 0.00). Models also tend to over-predict government-leaning stances, often misinterpreting ambiguous narratives. \ours and its associated diagnostics provide a foundation for advancing stance detection in Bangla media research and offer insights for improving LLM performance in low-resource languages.\footnote{\url{https://nusrat-lia.github.io/BanglaBias/}}

\end{abstract}

\section{Introduction}
\begin{figure}[h!]
\centering
\includegraphics[width=1\linewidth]{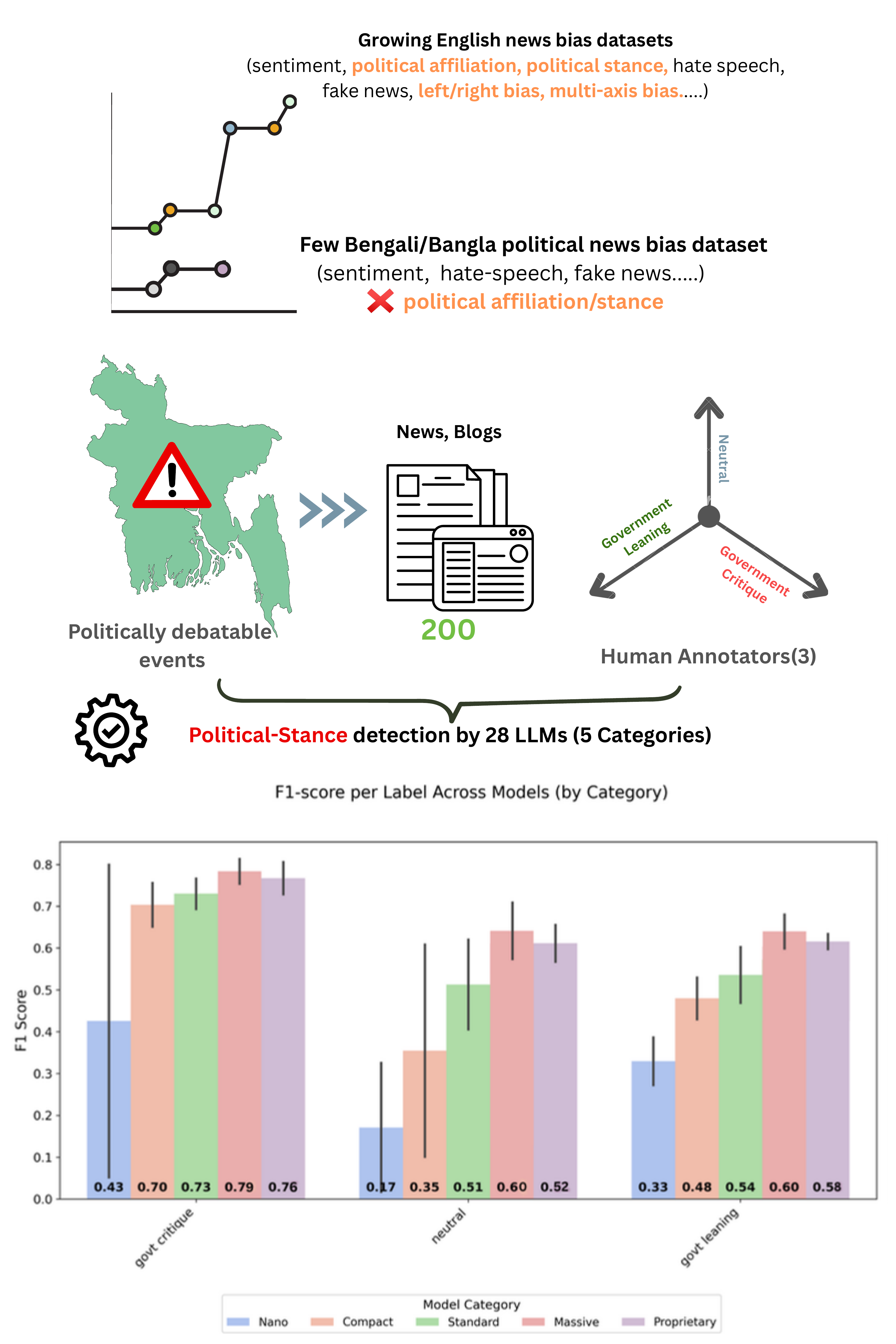}
\caption{Overview of political stance detection study (\textbf{Growing Resources for English vs. Lack of Bangla Resource Availability}). We introduce a benchmark of 200 news articles (on politically debatable events) annotated into Government Leaning, Critique, and Neutral labels. We then evaluate performance of 28 LLMs in detecting political stance in Bengali. Performance improves with model size, with Massive and Proprietary models achieving highest F1-scores, but neutral detection remains weak. Bars for Nano models and Neutral label show \textbf{noticeably larger error ranges across models, indicating unstable performance}.}
\vspace{-4.5mm}

\label{fig:intro}
\end{figure}

News media often reflect subtle political framing and bias, shaping public opinion and behavior \citep{fuhat2024mapping}. 
While extensive research has addressed this problem in English by creating rich datasets for sentiment analysis \cite{elbouanani2025analyzing, abercrombie2020parlvote}, stance detection \cite{rostami2025politisky24, khiabani2025cross}, hate speech identification \cite{hateoffensive}, and political bias classification \cite{cajcodes_political_bias}, the same cannot be said for many other languages. As \cref{fig:intro} illustrates, Bangla news media research faces a critical gap: despite clear evidence of political framing in coverage \cite{islam2016coverage}, there exists a scarcity of annotated datasets specifically designed to detect political stance and government affiliation bias. Our evaluations in \cref{fig:intro} revealed that while massive or proprietary models achieve strong performance on critique stance detection, neutral stance remains challenging, and Nano models showing large error bars underscores unstable performance in low-resource scenarios.

This disparity is particularly concerning given the widespread nature of political bias in Bangla news. Existing studies (both qualitative and computational) provide clear evidence that Bangla news content exhibits partisan frames that favor ruling-party narratives \citep{islam2016coverage, al2024investigating}. For instance, two news outlets covering the same political event, such as a government policy announcement, may frame it as either a progressive reform or an authoritarian overreach, despite reporting the same factual details \cite{vallejo2023connecting}. Such framing differences are not merely stylistic; they represent systematic \textit{ideological bias} -- a partisan slant away from neutral reporting \citep{mcquail2020mcquail}.

Computational studies have begun to tackle this with machine learning, e.g., Bangla BERT \cite{bhattacharjee2021banglabert} for hyper-partisanship \cite{mehadi2025bangla} and innovative LLM credibility bias \cite{prama2025evaluating}. However, these efforts are hampered by the lack of comprehensive and well-annotated datasets that can evaluate robust detection of political positions, specifically for the Bangladeshi socio-political landscape. Audience studies further highlight how perceived slant and censorship are reshaping Bangla news consumption \cite{islam2025conventional}.

To address this gap, we begin with collecting politically debatable events sourced from diverse news outlets and blogs. We then curated a dataset of 200 samples annotated by three native speakers for government stance (Government Leaning, Government Critique, or Neutral). \ours addresses the challenges of detecting political stance in Bangla news, as this is far more challenging than in English due to limited prior corpora, nuanced political language, frequent code-switching with English terms, and context-dependent rhetorical styles that make bias subtle and culturally embedded. We then evaluate reasoning-based political stance detection across 28 large language models (proprietary and open-source) to assess their effectiveness for this task.

To summarize, our work revolves around three key contributions:

\begin{itemize}[leftmargin=16pt, itemsep=0pt]

\item We introduce the first-ever benchmark dataset for political stance detection in Bangla with comprehensive metadata and a leaderboard. The dataset provides contextual information to study how political narratives are constructed in Bangladeshi media and enables the development and evaluation of stance detection models in such a low-resource setting.

\item We demonstrate multiple LLMs' effectiveness in detecting political bias in low-resource settings, with larger models achieving strong performance on Government Critique (F1=0.78) and Neutral (F1=0.61) classifications. 

\item We provide a systematic analysis that reveals important performance differences between different sizes of models as well as their biases and error tendencies.

\end{itemize}

\section{Relevant Work}

With the growing body of research on framing in NLP \citep{card2015media, fan2019plain, baly2020we, ziems2021protect}, and LLM prompting strategies \citep{brown2020language,achiam2023gpt, touvron2023llama} to detect stances, recent studies have explored media bias through multiple computational approaches, as we discuss in the following sections.

\subsection{Framing \& Political Bias in News Media}
\citet{entman2007framing} defines framing as the process of shaping narratives to promote specific interpretations, which then primes audiences to think in particular ways. Computational NLP work has built on these ideas by annotating and classifying news frames and biases \citep{card2015media,guida2025retain}. Early efforts include the Media Frames Corpus \citep{card2015media} and its multilingual extension \citep{piskorski2023semeval}, while others focus on ideological bias, such as BASIL \citep{fan2019plain}, AllSides \citep{baly2020we}, and smaller resources on regional or issue-specific perspectives \citep{lin2006side, chen2018learning, ziems2021protect}. Such datasets do not capture the socio-political nuances of Bangladeshi cultural context, and there exists no such Bengali counterpart yet.

In practice, most NLP work treats media bias as a classification problem: one common approach is single-label classification of an article's ideology. For instance, \citet{recasens2013linguistic} derived word-level ideological features; \citet{spinde2021identification} identified biased words via embedding distances between left- and right-leaning. 
Transformer-based models (e.g., BERT variants) now dominate framing tasks \citep{liu2019detecting,akyurek2020multi,piskorski2023semeval}. \citet{mehadi2025bangla} presented a Bangla BERT model fine-tuned to identify hyperpartisan (extremely biased) news articles, achieving 95.7\% accuracy, using a semi-supervised approach. The authors emphasized Bangla's ``low-resource''  status -- few annotated corpora, lexicons or pretrained models, and therefore relied on clever workarounds, i.e., machine translation (MT) for data augmentation. While MT is a practical approach, it fails to identify the socio-political facts that are specific to the region.

\subsection{LLMs for Political \& Subjective Analysis}
Large pretrained language models have become popular for political stance and bias tasks via prompt or instruction-based methods. Researchers have evaluated models like GPT-3/GPT-4 and open alternatives (Falcon, LLaMA, Mistral, etc.) on stance detection and bias classification \citep{faulborn2025only, ng2025examining}. For example, \citet{ng2025examining} tested GPT-3.5 and seven open-source LLMs on three stance datasets (SemEval-2016 tweets \citep{mohammad2016semeval}, Elections-2016 \citep{sobhani2017dataset} tweets, and the BASIL \citep{fan2019plain} news articles); they tried different prompting schemes (task description, context, chain-of-thought) and found reasoning based prompting scheme demonstrating the best performance in stance classification.

\citet{sucu2025exploiting} showed that adding user-context information to prompts can boost performance: in a political forum stance task, contextual prompting improved accuracy by 17.5\% to 38.5\% over baseline. \citet{lihammer2023semantic} showed GPT-3.5 and GPT-4 being able to reproduce political viewpoints. When tested in Bangladeshi context, most LLMs favored the left-leaning  sources, giving higher credibility ratings \citep{prama2025evaluating}. Instruction-tuned LMs can perform stance/bias tasks in a zero-shot or few-shot manner, but their outputs must be carefully validated as they may favor certain viewpoints and result in AI-induced bias in evaluating news sources, more significantly in low-resource languages like Bangla \citep{ng2025examining, prama2025evaluating}.

\section{Creating \ours}
Prior research on Bangla social discourse classification \citep{haider2024banth} highlighted the challenges of transliteration noise, subtle category distinctions, and embedding reliability. Beyond these technical difficulties, a major bottleneck remains -- the scarcity of high-quality, publicly available, politically nuanced Bangla datasets, which limits the development and evaluation of models for political stance detection and bias analysis. While translation or synthetic dataset generation has shown promise in domains like mathematics \cite{toshniwal2024openmathinstruct}, instruction following \cite{kim2024instatrans}, or general language tasks \cite{liu2024translation}, these approaches are often ineffective for political stance detection and media bias analysis in Bangla as socio-political contexts are deeply tied to cultural, historical, and linguistic nuances that are lost or distorted in translation. Building on the research gap, we developed a systematic pipeline to create the first annotated Bangla political stance detection  benchmark dataset.

For collecting the articles and human annotations, we followed these steps: 

\paragraph{Event Selection and Categorization:}
\label{sec:event_section}
We identified 46 socio-political events (full list in \cref{sec:events}) that generated significant public controversy or divergent viewpoints across the political spectrum. Our selection criteria prioritized events that: (1) received substantial media coverage across multiple outlets, (2) elicited clear government support or criticism in public discourse, and (3) represented diverse policy domains to ensure broad coverage of political stance patterns.

The temporal distribution spans Bangladesh's political landscape during the years 2013--2025. The early years (2013-2015) witnessed industrial disasters, the Shahbagh and Hefazat movements, and the contested 2014 election. The middle phase (2016-2019) was marked by the Holey Artisan attack, the Rohingya refugee influx, student protests on quota reform and road safety, and the 2018 election. The pandemic years (2020-2021) brought COVID-19, garment wage unrest, anti-rape protests, and growing dissent. The recent phase (2022-2025) includes fuel price hikes, the Padma Bridge opening, Rampal power plant launches, BNP–Awami League mega rallies, the boycotted 12th election, and violent quota protests with internet shutdowns and curfews.

\paragraph{Crawling:} We implemented a crawler to collect news articles for each event title based on the year of occurrence and saved results in HTML format. If no relevant articles were retrieved for the target year, the search was extended to the following 5-6 years to capture delayed or retrospective coverage.  

\paragraph{Parsing and normalization:} The HTML files were parsed and normalized to retain only the relevant data points, excluding advertisements, navigation menus, and unrelated web content.  

\paragraph{Article selection:} For each event, at least one and up to five representative articles were selected. We prioritized articles from established mainstream outlets including Prothom Alo, The Daily Star (Bangla), Jugantor, and Kaler Kantho, alongside politically diverse sources spanning the ideological spectrum. 

\paragraph{Human Annotation:} News items were annotated for political stance using a three-category framework: \textit{Government Leaning} (content that supports or favors government positions), \textit{Government Critique} (content that criticizes government actions), and \textit{Neutral} (balanced reporting without clear partisan positioning). We focused on identifying implicit stance indicators beyond explicit political statements, including source selection, framing choices, and political landscape analysis of each article's timeline and contextual emphasis.

Two annotators performed the initial annotation in parallel. Disagreements between these two annotators were resolved by a third annotator (Annotator 3), who acted as the adjudicator; the adjudicated labels are reported as final label (see annotator information in \cref{sec:annotator_info}).

Annotator 1 and Annotator 2 achieved 73.5\% agreement (Cohen's $\kappa$ = 0.574), indicating moderate inter-rater reliability. Most of the disagreements arose from neutral vs government critique and some neutral vs government leaning confusions. Disagreements were resolved by a third annotator (adjudicator) after mutual discussion and reanalysis of the political affiliations of the respective news media. Annotators primarily adhered to the stance detection decision flow outlined in \cref{sec:annotation-dec-flow}.

Our event-driven design captures politically significant moments where public opinion and media polarization are most visible, considering nuanced stance signals across politically diverse outlets. This also ensures quality news dataset while decreasing the chances of noisy data. \textbf{Our news collection pipeline is reusable and extensible to any low- or high-resource language}. This allows future researchers to expand the dataset with new events or apply it to other political contexts.

\subsection{Dataset Statistics}
\label{sec:data_stats}
\begin{figure}[!t] 
\centering
\includegraphics[width=\linewidth]{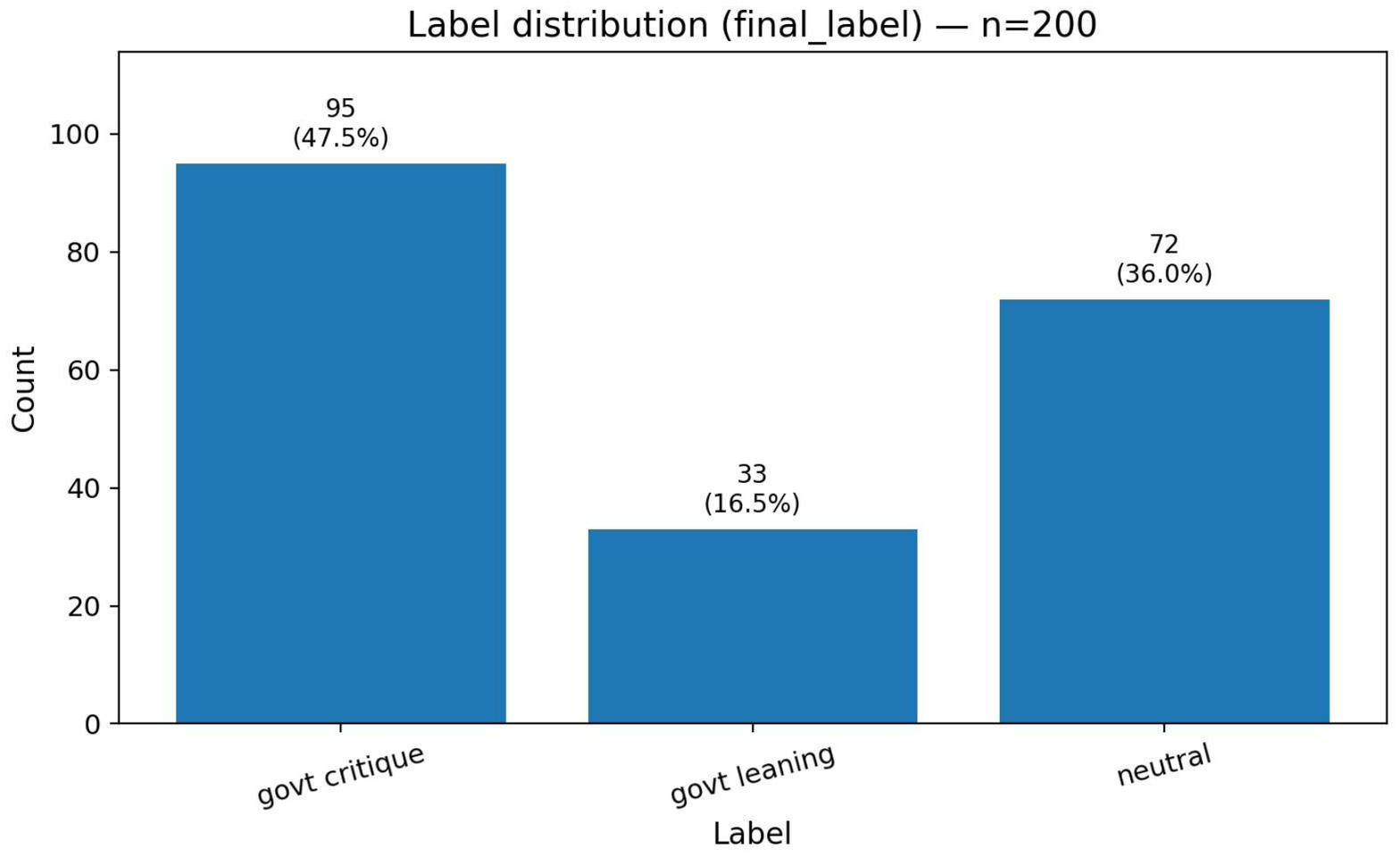}
\caption{Distribution of three classes across the dataset: 95 Govt. Critique (47.5\%), 72 Neutral (36.0\%), and 33 Govt. Leaning (16.5\%).}
\label{fig:stance_class_distribution}
\end{figure}

In total there are 200 data samples and 8 key information; every item has a unique \textit{ID}, a \textit{News Body}, a \textit{Headline}, a \textit{Source link} and the \textit{Date}. The \textit{Event} field contains 46 distinct events, and \textit{News Corpora Name} lists 54 distinct sources with the largest contributors being BBC Bangla (N=37), Prothom Alo (N=20), Jugantor (N=17), DW (N=11) and Bangla Tribune (N=11) (full details reported in \cref{tab:news_corpora_distribution}). As \cref{fig:stance_class_distribution} presents, the final label has three classes (govt. critique: 95, Neutral: 72, govt. leaning: 33). Because nearly half the dataset is labeled ``govt. critique'' while ``govt. leaning'' is underrepresented ($\approx$16.5\%), we recommend treating class imbalance explicitly during evaluation.

\section{Experimental Setup}
We benchmark the performance of several state-of-the-art large language models (LLMs) on the political stance detection using our benchmark dataset of 200 Bangla articles annotated with one of three stances: Government Leaning, Government Critique, or Neutral.

\subsection{Evaluation Metrics}
\begin{table*}[t]
\centering
\resizebox{\textwidth}{!}{
\begin{tabular}{@{}llcccccccccccc@{}}
\toprule
 &  & \multicolumn{3}{c}{\textbf{\textit{Govt. Critique}}} & \multicolumn{3}{c}{\textbf{\textit{Neutral}}} & \multicolumn{3}{c}{\textbf{\textit{Govt. Leaning}}} & \multicolumn{3}{c}{\textbf{\textit{Weighted Avg.}}} \\
\cmidrule(lr){3-5}  \cmidrule(lr){6-8}  \cmidrule(lr){9-11}  \cmidrule(lr){12-14} 
 &  & P & R & F1 & P & R & F1 & P & R & F1 & P & R & F1 \\
\midrule
\multirow[t]{2}{*}{Nano}  & Qwen3-1.7B & 0.71 & 0.69 & 0.70 & 0.54 & 0.21 & 0.30 & 0.28 & 0.67 & 0.39 & 0.58 & 0.52 & 0.51 \\
 & Qwen3-0.6B & 0.58 & 0.57 & 0.57 & 0.35 & 0.15 & 0.21 & 0.22 & 0.52 & 0.31 & 0.44 & 0.41 & 0.40 \\
 \cmidrule{1-14}
\multirow[t]{2}{*}{Compact}  & TigerLLM-9B & 0.74 & 0.78 & 0.76 & 0.62 & 0.65 & 0.64 & 0.62 & 0.45 & 0.53 & 0.68 & 0.68 & 0.68 \\
 & Qwen3-8B & 0.75 & 0.73 & 0.74 & 0.62 & 0.42 & 0.50 & 0.42 & 0.76 & 0.54 & 0.65 & 0.62 & 0.62 \\
 \cmidrule{1-14}
\multirow[t]{2}{*}{Standard}  & Qwen3-32B & 0.82 & 0.68 & 0.75 & 0.55 & 0.72 & 0.63 & 0.70 & 0.58 & 0.63 & 0.71 & 0.68 & 0.68 \\
 & GPT-OSS-20B & 0.75 & 0.72 & 0.73 & 0.55 & 0.60 & 0.57 & 0.55 & 0.52 & 0.53 & 0.64 & 0.64 & 0.64 \\
\cmidrule{1-14}
\multirow[t]{2}{*}{Massive}  & Qwen3-235B-I & 0.78 & 0.86 & 0.82 & 0.71 & 0.62 & 0.67 & 0.66 & 0.64 & 0.65 & 0.74 & 0.74 & 0.74 \\
 & Llama3.3-70B & 0.77 & 0.91 & 0.83 & 0.80 & 0.50 & 0.62 & 0.57 & 0.76 & 0.65 & 0.75 & 0.73 & 0.73 \\
\cmidrule{1-14}
\multirow[t]{2}{*}{Proprietary}  & Claude-Sonnet-4 & 0.90 & 0.73 & 0.80 & 0.63 & 0.69 & 0.66 & 0.52 & 0.70 & 0.60 & 0.74 & 0.71 & 0.72 \\
 & Gemini-2.5-Pro & 0.79 & 0.77 & 0.78 & 0.61 & 0.61 & 0.61 & 0.56 & 0.61 & 0.58 & 0.69 & 0.69 & 0.69 \\

\bottomrule
\end{tabular}
}
\caption{Performance metrics (Precision, Recall, F1) for the top two models per category, ranked by weighted F1.}
\label{tab:top2_model_performance}
\end{table*}

Our primary evaluation frames political stance detection as a \textsc{Ternary Classification} task. Each data point consists of an article text and a ground-truth label (Government Leaning, Neutral, or Government Critique). Given the article, models must predict the stance. We evaluate models on this task using standard metrics including accuracy, precision, recall, and weighted F1 score.

We conduct evaluations in a reasoning-based detection setting. Following \citet{ng2025examining}, we prompted models to classify the article and provide a brief reasoning. This framework supports additional qualitative analyses to reveal performance differences across model sizes:

\begin{enumerate}[label=(\arabic*),itemsep=0pt]
\item \textbf{Per-Label Performance} assesses F1 scores for each stance category to identify imbalances in handling Government Leaning, Neutral, or Government Critique articles.
\item \textbf{Confusion Analysis} uses confusion matrices (aggregated by model category) to highlight common misclassification patterns, such as confusing Neutral with Government Leaning.
\item \textbf{Bias Tendency} evaluates prediction distributions via radar plots, comparing them to the ground-truth distribution to detect systematic biases toward certain stances.
\item \textbf{Misclassification Analysis} examines incorrect predictions alongside model-generated reasons to uncover qualitative error patterns, such as over-reliance on specific keywords or failure to detect nuance.

\end{enumerate}

\subsection{Models}
\label{sec:models}
We evaluate a diverse set of 28 language models, categorized by parameter size: Nano ($<$2B parameters), Compact (2B--10B), Standard (10B--40B), Massive ($>$40B), and Proprietary (size undisclosed). This selection includes both open-source and closed-source models to ensure comprehensive coverage.

\begin{itemize}[itemsep=0pt]
    \item \textbf{Nano}: Qwen2.5-0.5B, Qwen3-0.6B, Qwen3-1.7B.
    
    \item \textbf{Compact}:  Qwen2.5-3B, Qwen3-4B, Qwen3-4B-I, Qwen3-4B-T, Qwen3-8B, Llama3.1-8B, TigerLLM-9B.
    
    \item \textbf{Standard}: Qwen2.5-14B, Qwen3-14B, Qwen3-30B-I, Qwen3-30B-T, Qwen3-32B, GPT-OSS-20B.
    
    \item \textbf{Massive}: Llama3.3-70B, GPT-OSS-120B, Qwen3-235B-I, Qwen3-235B-T, GLM-4.5, DeepSeek-V3.1, DeepSeek-R1.
    
    \item \textbf{Proprietary}: Claude-Sonnet-4, Grok-4, Gemini-2.5-Pro, GPT-4.1-Mini, GPT-5-Mini.
\end{itemize}

Models are prompted to generate both a stance label (Decision: D) and reasoning (R) (R$\to$D).
Full model ids can be found in \cref{sec:model_name_map}. The evaluated models include families such as Qwen \cite{bai2023qwen}, LLaMA \cite{touvron2023llama}, GLM \cite{zeng2025glm}, DeepSeek \cite{bi2024deepseek}, GPT \cite{achiam2023gpt}, Claude \cite{caruccio2024claude}, and Gemini \cite{ng2025examining}.

\section{Results \& Analysis}
\label{sec:results_analysis}

\begin{figure*}[!ht]
    \centering
    \includegraphics[width=\textwidth]{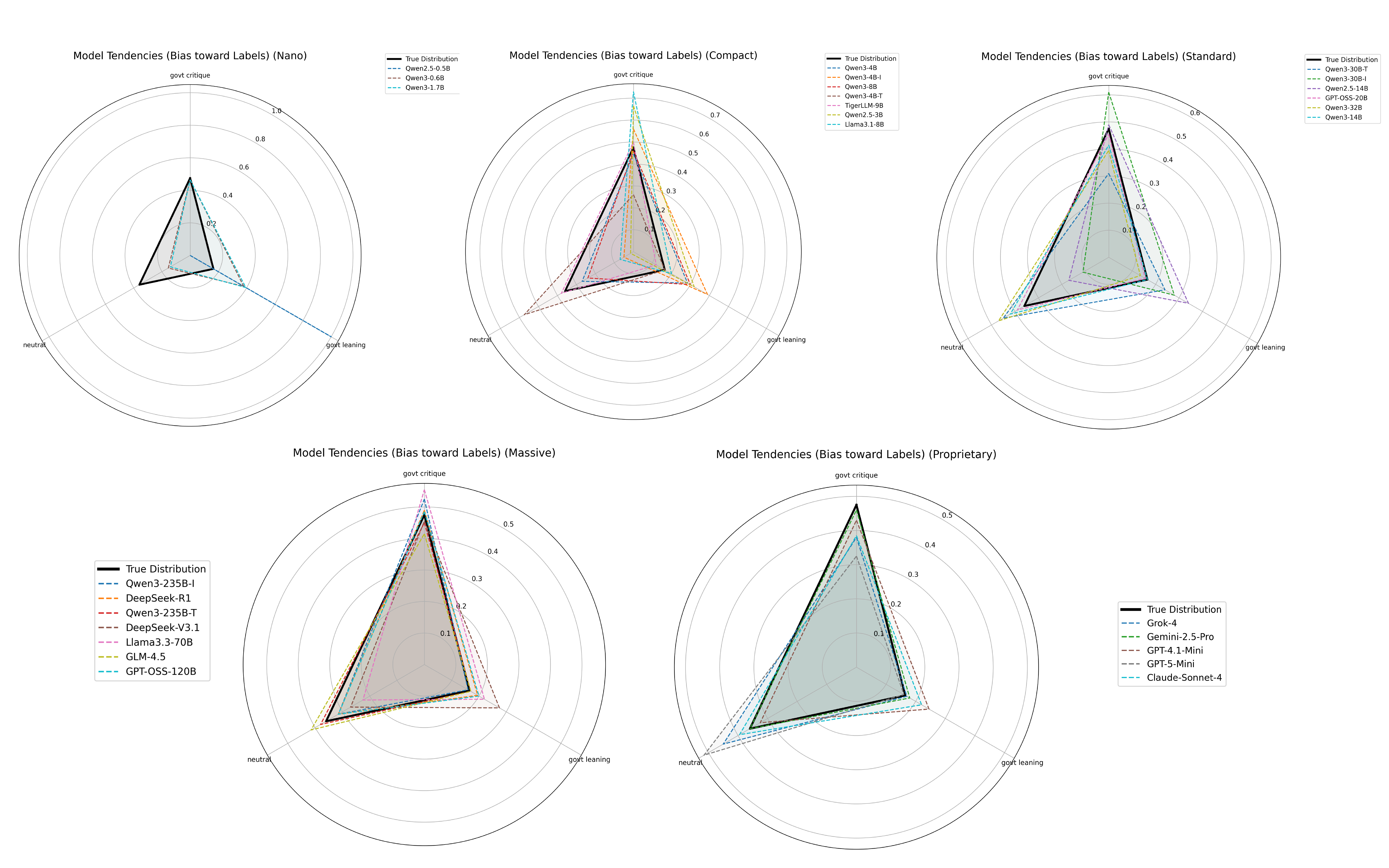}
    \caption{Radar plots showing tendencies of models (per-category) to favor particular labels, relative to the true distribution. The black polygon in each radar plot denotes the true distribution of labels and serves as the baseline.}
    \label{fig:bias_tendency}
\end{figure*}

\subsection{Performance Metrics Analysis}
\cref{tab:top2_model_performance} presents the performance of the top two models from each category (ranked by weighted average F1 score) in political stance detection task. Results for all 28 models can be found in \cref{tab:all_model_performance_table}.

\finding{\textbf{Finding 1}: Neutral Category Exposes Model Weaknesses}
Most models struggle with neutral content, with F1 scores as low as 0.00 for Qwen2.5-0.5B and 0.16 for Llama3.1-8B due to poor recall (e.g., 0.10 for Llama3.1-8B). Even massive models like Qwen3-235B-T only reach 0.68 (F1). The consistent struggle across model sizes points to potential under-representation of Bangla neutral samples and insufficient learning of contextual cues that distinguish neutral stances from biased ones. The challenge is particularly evident in articles requiring balanced interpretation of government actions without explicit sentiment, indicating a gap in models' ability to handle nuanced, non-polarized Bangla narratives (see \cref{tab:all_model_performance_table}).

\finding{\textbf{Finding 2}: Government Critique Stance is the Easiest One to Classify}
Models excel at detecting government critique stance, with top performers Llama3.3-70B and Qwen3-235B-I achieving F1 scores of 0.83 and 0.82. We hypothesize that, critical contents contain distinct linguistic cues such as strong negative sentiment and keywords (e.g., Government negligence, fraud, Repressive policy, Negligence). This strength highlights a potential bias in model design, as the same models struggle with neutral content, suggesting an over-reliance on sentiment cues rather than nuanced contextual understanding.

\finding{\textbf{Finding 3}: Bias Towards Government-Leaning Stance}
In government-leaning stance prediction, massive models like Qwen3-235B-I and Llama3.3-70B achieve decent F1 scores, but recall often exceeds precision (\cref{tab:all_model_performance_table}). This indicates a tendency to over-predict supportive stances. This imbalance suggests that models are overly sensitive to cues that might indicate government support, leading to false positives. This over-prediction risks amplifying perceived government support in media analysis, potentially skewing automated content moderation or sentiment analysis applications.

\finding{\textbf{Finding 4}: Smaller Models can be Efficient (to some extent)}
Although large-scale models (Massive and Proprietary) demonstrate clear dominance in classification tasks, standard models like Qwen3-32B and even smaller TigerLLM \cite{raihan2025tigerllm} show competitive results in detecting stances, particularly government critique and neutral. This efficiency suggests that smaller models can effectively handle key linguistic features, such as critical keywords or neutral tone indicators, without requiring the computational resources of their larger counterparts. The strong performance of TigerLLM, in particular, highlights the potential of small domain-optimized models tailored for Bangla text, which benefit from focused training on regional linguistic patterns. In low-resource settings, where computational constraints are a significant concern, such resource-efficient models can be further optimized to perform critical stance detection tasks. 

\subsection{Bias Tendency}

\begin{figure*}[!ht]
    \centering
    \includegraphics[width=\textwidth]{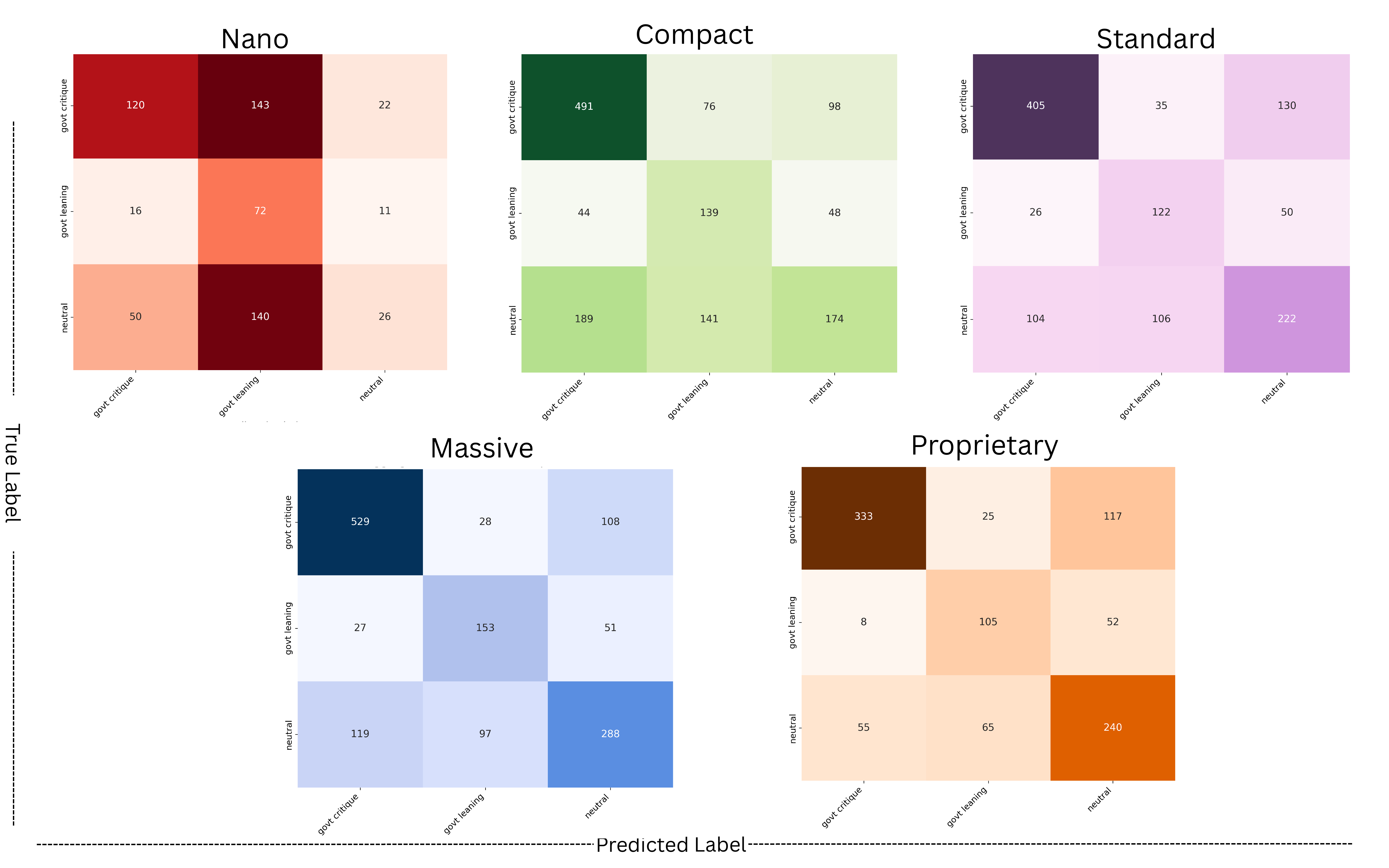}
    \caption{Aggregated Confusion Heatmap over five categories of models: Nano (3 models), Compact (7 models), Standard (6 models), Massive (7 models) and Proprietary (5 models).}
    \label{fig:confusion_matrix}
\end{figure*}

\cref{fig:bias_tendency} illustrates models' tendencies across Nano, Compact, Standard, Massive, and Proprietary categories to favor particular labels, depicting biases in understanding Bangla linguistic variations. Nano models like Qwen2.5-0.5B, Qwen3-1.7B display extreme bias toward government leaning stance. Compact models vary in biases, with TigerLLM-9B showing decent alignment. Qwen3-4B-T leans towards neutral more often, whereas Llama3.1-8B mostly mistakes neutral articles as critique stances. Standard models like Qwen2.5-14B show bias towards government leaning stance prediction, while Qwen3-30B-I struggles with over-predicting critique label. Massive models including Llama3.3-70B and Qwen3-235B-I cluster closer to the true distribution with minimal deviations. Proprietary models like Gemini-2.5-Pro and GPT-4-Mini align well overall, but GPT-5-mini leans more towards neutral label.

\vspace{-1.2mm}

\paragraph{But Where the Bias Lies?}
The confusion heatmaps in \cref{fig:confusion_matrix} reveal that smaller model categories like Nano and Compact frequently confuse neutral labels with government-leaning or critique, as seen in Nano's 140 neutral instances misclassified as government-leaning and Compact's 189 neutrals mistaken for government-critique. This indicates a bias toward polarized predictions due to limited capacity. As model scale increases, such as in Standard and Massive categories, mispredictions decrease, with Massive models showing balanced performance but still confusing neutral with government-critique. Proprietary models minimize cross-polarity errors (only 25 government-critique predicted as government leaning), yet persist in confusing neutral with polar stances (55 and 65 instances).

\subsection{Prompting Strategy Comparison:}
To further investigate the impact of prompting on stance detection, we compare zero-shot (no example) and few-shot (multiple examples) prompting strategies across selected models, focusing on those with varying sizes (e.g., 1.7B to 70B parameters). 

In subjective tasks like political bias classification in Bangla news articles, few-shot prompting can underperform compared to zero-shot, particularly in larger models (see \cref{sec:prompting-comparison}), due to risks of overfitting to example patterns, such as over-emphasizing neutrality if examples are imbalanced or create conservative heuristics that dismiss implicit biases as ``factual.'' This leads to systematic errors like defaulting to ``Neutral'' labels, as seen in aggregate metrics where few-shot Weighted F1 scores lag (e.g., 0.57--0.67 vs. zero-shot's 0.68--0.73). However, few-shot may still edge out in smaller models (e.g., 1.7B) needing more examples for calibration. This highlights that prompting-strategy efficacy depends on model scale, multiple patterns of framing and dataset nuances like implicit political tones.

\subsection{Error Analysis}
\label{sec:error_analysis}

\cref{tab:error_analysis_table} shows some example error cases with the LLMs' reasoning. Error analysis of model predictions reveals three main failure modes. 

\paragraph{Content-balance Ambiguity:} Many misclassified articles are fact-based, presenting both government and opposition, or including technical detail that lacks an evaluative authorial stance. Such articles require understanding of external political events. For such items, the ground-truth labels often force a single polar stance, but models predict neutral as they do not detect any favoring language. 

In such instances, models often reason that the article presents both parties and hence they conclude it to be neutral. Such reasoning misses the polar stance as it fails to incorporate understanding of any external entity or situational complexity.

For example, as shown in \cref{tab:error_analysis_table}, in Article ID 110 (predicted neutral by Qwen3-32B, true label: govt. critique), the model reasoned that the article reports conflicting claims between protesters and the government-aligned student organization without taking a clear stance, presenting both accusations and denials. However, this overlooks the subtle emphasis on critiques through the framing of events, and shows how balanced presentation can mask underlying bias when external context (e.g., ongoing political tensions) is not considered.

\paragraph{Lexical-cue Over-reliance:} When articles contain explicit praise (honorifics, laudatory framing), models  predict government leaning stance even if the article was neutral. This stems from the fact that in Bangla articles, the mention of any leader name often comes with a laudatory term (``Respected Minister''), even though the articles are factual. Articles praising government initiatives when the actions were actually praise worthy, were classified as government leaning stance, despite it being neutral. 

We hypothesize that as the article praises such persona, it must mean a favoring stance. Predicting such article stance is much harder as it requires world knowledge and context outside the immediate text \cite{burnham2025stance}.

Examples include Article 98 (predicted govt. leaning by DeepSeek-V3.1, true label: neutral) (see \cref{tab:error_analysis_table}), where the focus on a government lockdown was seen as supportive due to proactive framing, ignoring the fact based neutral trait. Similarly, in Article 148 (predicted govt. leaning by Qwen3-235B-I, true label: neutral), models reasoning highlighted government quotes against BNP, ignoring the objective situational context.

\paragraph{Selective Perspective Emphasis:} A third common failure mode involves models amplifying one side's narrative due to disproportionate quoting or structural prominence, leading to wrong predictions in otherwise balanced articles. This occurs particularly in dynamic political contexts where opposition voices are detailed extensively, even if countered, causing models to infer critique stance.

Similarly, events framed around protests or right-abuses push models toward government critique stance prediction, despite largely factual reporting. These patterns suggest models' lack of understanding of speaker-aware representation and dependency on generic sentiment cues.

An illustrative case is Article 25 (predicted govt. critique by Qwen3-4B-T, true label: neutral) (see \cref{tab:error_analysis_table}, where the reasoning emphasized the opposition's (BNP) detailed accusations against the government, including human rights violations, despite the news being just a factual representation of opinions. The model failed to recognize the fact based patterns, treating vivid critical language as dominant bias.

\section{Conclusion}
Detecting political stance in Bangla news media requires nuanced understanding of linguistic cues, cultural context, and subtle biases embedded in reporting. This paper introduces a novel benchmark dataset, \ours, designed to evaluate the political stance detection capabilities of computational approaches in media bias research, which can be extended to any low-resource setting. Our findings reveal that while most LLMs excel at identifying government-critical content, they struggle significantly with neutral content due to inadequate handling of context-dependent rhetorical styles, pointing to an under-representation of extensive samples in models' training data. Our analyses indicate a tendency to misinterpret ambiguous content as government-leaning. Significant performance gains by domain-optimized LLMs further fortify our findings on the need for high-quality, culturally grounded datasets and targeted fine-tuning to improve stance detection in low-resource, politically nuanced contexts. Extending \ours  framework to other underrepresented languages could further democratize computational media analysis and sets the stage for more equitable, accurate, and culturally informed bias detection systems globally.

\section*{Limitations}
Our study evaluates a diverse set of 28 large language models (LLMs) for political stance detection in Bangla news, covering both proprietary and open-source architectures. However, we acknowledge the rapid development of newer models or specialized architectures. We plan to extend our evaluation to incorporate emerging models and domain-specific adaptations to better capture Bangla-specific nuances.

\ours, while carefully curated to represent government-leaning, government-critique, and neutral stances, may not fully capture the diversity of Bangla media, particularly from regional or less prominent outlets. Additionally, the dataset relies on articles published before the models' training cutoffs, making it challenging to ensure that models have not encountered similar content during pre-training, which could inflate performance metrics. Future work will aim to expand the dataset with more diverse sources and explore methods to verify model exposure to training data.

The task of stance detection focused on textual content, specifically article bodies, without integrating multimodal elements such as images, news-videos, or social media metadata, which often influence framing in Bangla news. We leave the exploration of multimodal stance detection to future work.

Finally, due to computational constraints, we could not perform extensive few-shot experiments, particularly proprietary ones like Claude-Sonnet-4. For consistency, we focused on models with accessible APIs or open-source weights, but future work will explore fine-tuning strategies and broader few-shot learning to improve performance, especially for neutral content detection.

\bibliography{custom}
\clearpage

\appendix

\section{Performance metrics of all 28 Models}
\label{sec:all_model_performance}
\cref{tab:all_model_performance_table} shows stance detection performance evaluation results for all 28 models (5 categories).
\begin{table*}[!ht]
\centering
\resizebox{0.95\textwidth}{!}{
\begin{tabular}{@{}lccccccccccccc@{}}
\toprule
& \textbf{\textit{Accuracy}} & \multicolumn{3}{c}{\textbf{\textit{Govt. Critique}}} & \multicolumn{3}{c}{\textbf{\textit{Neutral}}} & \multicolumn{3}{c}{\textbf{\textit{Govt. Leaning}}} & \multicolumn{3}{c}{\textbf{\textit{Weighted Avg.}}} \\
\cmidrule(lr){3-5}  \cmidrule(lr){6-8}  \cmidrule(lr){9-11}  \cmidrule(lr){12-14} 
&  & P & R & F1 & P & R & F1 & P & R & F1 & P & R & F1 \vspace{1mm}\\
\midrule
\rowcolor{lightgray}\multicolumn{14}{l}{\it \textbf{Nano Models}} \\
Qwen2.5-0.5B & 0.17 & 0.00 & 0.00 & 0.00 & 0.00 & 0.00 & 0.00 & 0.17 & 1.00 & 0.28 & 0.03 & 0.17 & 0.05 \\
Qwen3-0.6B & 0.41 & 0.58 & 0.57 & 0.57 & 0.35 & 0.15 & 0.21 & 0.22 & 0.52 & 0.31 & 0.44 & 0.41 & 0.40 \\
Qwen3-1.7B & 0.52 & 0.71 & 0.69 & 0.70 & 0.54 & 0.21 & 0.30 & 0.28 & 0.67 & 0.39 & 0.58 & 0.52 & 0.51 \vspace{1mm}\\

\rowcolor{lightgray}\multicolumn{14}{l}{\it \textbf{Compact Models}} \\
Qwen2.5-3B & 0.50 & 0.59 & 0.83 & 0.69 & 0.00 & 0.00 & 0.00 & 0.33 & 0.64 & 0.43 & 0.34 & 0.50 & 0.40 \\
Qwen3-4B-I & 0.52 & 0.68 & 0.80 & 0.73 & 0.50 & 0.07 & 0.12 & 0.28 & 0.67 & 0.40 & 0.55 & 0.52 & 0.46 \\
Llama3.1-8B & 0.55 & 0.58 & 0.88 & 0.70 & 0.50 & 0.10 & 0.16 & 0.45 & 0.55 & 0.49 & 0.53 & 0.55 & 0.47 \\
Qwen3-4B-T & 0.57 & 0.85 & 0.46 & 0.60 & 0.48 & 0.76 & 0.59 & 0.48 & 0.48 & 0.48 & 0.65 & 0.57 & 0.58 \\
Qwen3-4B & 0.58 & 0.73 & 0.68 & 0.71 & 0.56 & 0.42 & 0.48 & 0.39 & 0.67 & 0.49 & 0.61 & 0.58 & 0.59 \\
Qwen3-8B & 0.62 & 0.75 & 0.73 & 0.74 & 0.62 & 0.42 & 0.50 & 0.42 & 0.76 & 0.54 & 0.65 & 0.62 & 0.62 \\
TigerLLM-9B & 0.68 & 0.74 & 0.78 & 0.76 & 0.62 & 0.65 & 0.64 & 0.62 & 0.45 & 0.53 & 0.68 & 0.68 & 0.68 \vspace{1mm}\\

\rowcolor{lightgray}\multicolumn{14}{l}{\it \textbf{Standard Models}} \\
Qwen2.5-14B & 0.57 & 0.72 & 0.75 & 0.74 & 0.62 & 0.29 & 0.40 & 0.34 & 0.70 & 0.46 & 0.62 & 0.57 & 0.57 \\
Qwen3-30B-I & 0.62 & 0.70 & 0.89 & 0.78 & 0.77 & 0.24 & 0.36 & 0.41 & 0.70 & 0.52 & 0.68 & 0.62 & 0.59 \\
Qwen3-14B & 0.61 & 0.76 & 0.66 & 0.71 & 0.51 & 0.60 & 0.55 & 0.48 & 0.48 & 0.48 & 0.62 & 0.61 & 0.61 \\
Qwen3-30B-T & 0.61 & 0.85 & 0.56 & 0.68 & 0.51 & 0.64 & 0.57 & 0.50 & 0.73 & 0.59 & 0.67 & 0.61 & 0.62 \\
GPT-OSS-20B & 0.64 & 0.75 & 0.72 & 0.73 & 0.55 & 0.60 & 0.57 & 0.55 & 0.52 & 0.53 & 0.64 & 0.64 & 0.64 \\
Qwen3-32B & 0.68 & 0.82 & 0.68 & 0.75 & 0.55 & 0.72 & 0.63 & 0.70 & 0.58 & 0.63 & 0.71 & 0.68 & 0.68 \vspace{1mm}\\

\rowcolor{lightgray}\multicolumn{14}{l}{\it \textbf{Massive Models}} \\
DeepSeek-V3.1 & 0.64 & 0.79 & 0.76 & 0.77 & 0.59 & 0.44 & 0.51 & 0.42 & 0.70 & 0.52 & 0.66 & 0.64 & 0.64 \\
DeepSeek-R1 & 0.67 & 0.76 & 0.78 & 0.77 & 0.60 & 0.53 & 0.56 & 0.54 & 0.64 & 0.58 & 0.66 & 0.67 & 0.66 \\
GPT-OSS-120B & 0.67 & 0.76 & 0.78 & 0.77 & 0.59 & 0.51 & 0.55 & 0.55 & 0.67 & 0.60 & 0.66 & 0.67 & 0.66 \\
GLM-4.5 & 0.69 & 0.82 & 0.72 & 0.76 & 0.60 & 0.69 & 0.65 & 0.59 & 0.61 & 0.60 & 0.70 & 0.69 & 0.69 \\
Qwen3-235B-T & 0.72 & 0.81 & 0.77 & 0.79 & 0.66 & 0.69 & 0.68 & 0.62 & 0.64 & 0.63 & 0.72 & 0.72 & 0.72 \\
Llama3.3-70B & 0.73 & 0.77 & 0.91 & 0.83 & 0.80 & 0.50 & 0.62 & 0.57 & 0.76 & 0.65 & 0.75 & 0.73 & 0.73 \\
Qwen3-235B-I & 0.74 & 0.78 & 0.86 & 0.82 & 0.71 & 0.62 & 0.67 & 0.66 & 0.64 & 0.65 & 0.74 & 0.74 & 0.74 \vspace{1mm}\\

\rowcolor{lightgray}\multicolumn{14}{l}{\it \textbf{Proprietary Models}} \\
GPT-5-Mini & 0.65 & 0.88 & 0.60 & 0.71 & 0.52 & 0.75 & 0.62 & 0.56 & 0.55 & 0.55 & 0.70 & 0.65 & 0.65 \\
GPT-4.1-Mini & 0.66 & 0.81 & 0.74 & 0.77 & 0.58 & 0.53 & 0.55 & 0.49 & 0.73 & 0.59 & 0.68 & 0.66 & 0.66 \\
Grok-4 & 0.69 & 0.84 & 0.67 & 0.75 & 0.60 & 0.75 & 0.67 & 0.59 & 0.61 & 0.60 & 0.71 & 0.69 & 0.69 \\
Gemini-2.5-Pro & 0.69 & 0.79 & 0.77 & 0.78 & 0.61 & 0.61 & 0.61 & 0.56 & 0.61 & 0.58 & 0.69 & 0.69 & 0.69 \\
Claude-Sonnet-4 & 0.71 & 0.90 & 0.73 & 0.80 & 0.63 & 0.69 & 0.66 & 0.52 & 0.70 & 0.60 & 0.74 & 0.71 & 0.72 \\
\bottomrule
\end{tabular}
}
\caption{Accuracy and label-wise performance metrics of models on \ours benchmark dataset. For each label (Government Critique, Neutral, Government Leaning), precision (P), recall (R), and F1-scores are reported, along with the overall weighted averages. For each block, the results are sorted based on the Weighted F1 score.}
\label{tab:all_model_performance_table}
\end{table*}

\clearpage

\section{Examples of Wrongly Predicted Label and Model-Reasoning}
\cref{tab:error_analysis_table} shows five error examples along with models' reasoning.
\label{sec:error_analysis_appen}
\begin{table*}[ht]
\centering
\resizebox{\textwidth}{!}{
\begin{tabular}{@{}llllp{8cm}@{}}
\toprule
\textbf{Article ID} & \textbf{Model Name} & \textbf{True Label} & \textbf{Predicted Label} & \textbf{Reasoning} \\
\midrule
20 & Claude-Sonnet-4 & Govt. Leaning & Neutral & The article provides factual reporting about Awami League's candidate announcement for the 12th national parliament election, including specific names and constituencies. It presents information objectively without editorial commentary or bias toward or against the government. \\
\midrule
25 & Qwen3-4B-T & Neutral & Govt. Critique & The article prominently features the opposition's (BNP) accusations against the government as 'unlawful, unelected, and occupying the state power' with detailed claims of human rights violations and violence. While it includes the government's response, the narrative consistently emphasizes the opposition's critique of the government's actions and policies.\\
\midrule
98 & DeepSeek-V3.1 & Neutral & Govt. Leaning & The article reports the government's decision to impose a strict lockdown in a factual manner, focusing on the official announcement and the measures being taken. It presents the government as actively responding to the crisis without including significant critical perspectives or questioning the decision. \\
\midrule
110 & Qwen3-32B & Govt. Critique & Neutral & The article reports conflicting claims between protesters and the government-aligned student organization without taking a clear stance. It presents both accusations and denials objectively, focusing on factual events without evident bias.\\
\midrule
148 & Qwen3-235B-I & Neutral & Govt. Leaning & The article predominantly presents the government's perspective, portraying BNP's demands as unreasonable and disruptive, while highlighting state efforts to maintain peace and order. Government officials' statements are quoted extensively to justify decisions and frame BNP's actions as potentially inciting chaos. \\

\bottomrule
\end{tabular}
}
\caption{Selected cases of misclassified articles in our Bangla political stance detection study. Each row lists the article ID, the model used, the true label, the predicted label, and the reasoning provided by the model.}
\label{tab:error_analysis_table}
\end{table*}

\clearpage

\section{News Sources and Respective Article Count}
\cref{tab:news_corpora_distribution} includes all the news sources and the number of articles sourced from respective corpora, blog or site.
\label{sec:news_source}
\begin{table*}[ht]
\centering
\begin{tabular}{@{}lc lc@{}}

\toprule
\textbf{News Source} & \textbf{Article Count} & \textbf{News Source} & \textbf{Article Count} \\
\midrule
BBC Bangla & 37 & Doinik Bangla & 1 \\
Prothom Alo & 20 & Bagerhatinfo & 1 \\
Jugantor & 17 & USEmbassy & 1 \\
DW & 11 & Dainik Shiksha & 1 \\
Bangla Tribune & 11 & Bonik Barta & 1 \\
The Daily Star & 10 & Bangladesh Jamate Islam & 1 \\
Samakal & 7 & Kaler Kontho & 1 \\
Bdnews24 & 7 & Bdnew24 & 1 \\
Dhaka Post & 7 & News Bangla & 1 \\
Bangla News24 & 7 & Bangla 24 live newspaper & 1 \\
Jago News & 6 & Dhaka Times  & 1 \\
Somoy news  & 5 & SattAcademy & 1 \\
Dhaka Times & 5 & voabangla & 1 \\
Daily Ittefaq & 4 & Daily Janakantho & 1 \\
Daily Inqilab & 4 & Ajker Potrika & 1 \\
Dhaka Tribune & 2 & Desh Rupantor & 1 \\
Somoyer Alo & 2 & Daily Campus & 1 \\
Channel online & 2 & Khulna Gazet & 1 \\
The Daily Vorer Pata & 1 & Cvoice24 & 1 \\
Ajkaler khobor & 1 & Dainik Sylhet & 1 \\
BanglaTribune & 1 & Dainik amader bangla & 1 \\
smsaif & 1 & Chalaman neywork & 1 \\
somewhereinblog & 1 & The doctors dialogue & 1 \\
Timetouch news & 1 & Kaler kontho & 1 \\
Banik Barta & 1 & somewhere in blog & 1 \\
DBC news & 1 & rajibkhaja & 1 \\
MuktiBarta & 1 & Green Watch BD & 1 \\
\bottomrule
\end{tabular}
\caption{Distribution of articles across news sources, showing the number of collected articles from each outlet.}
\label{tab:news_corpora_distribution}
\end{table*}

\clearpage
\section{Politically Debated Events Covered by \ours}
\label{sec:events}
\begin{figure*}[ht]
    \centering
    
    \includegraphics[width=\textwidth]{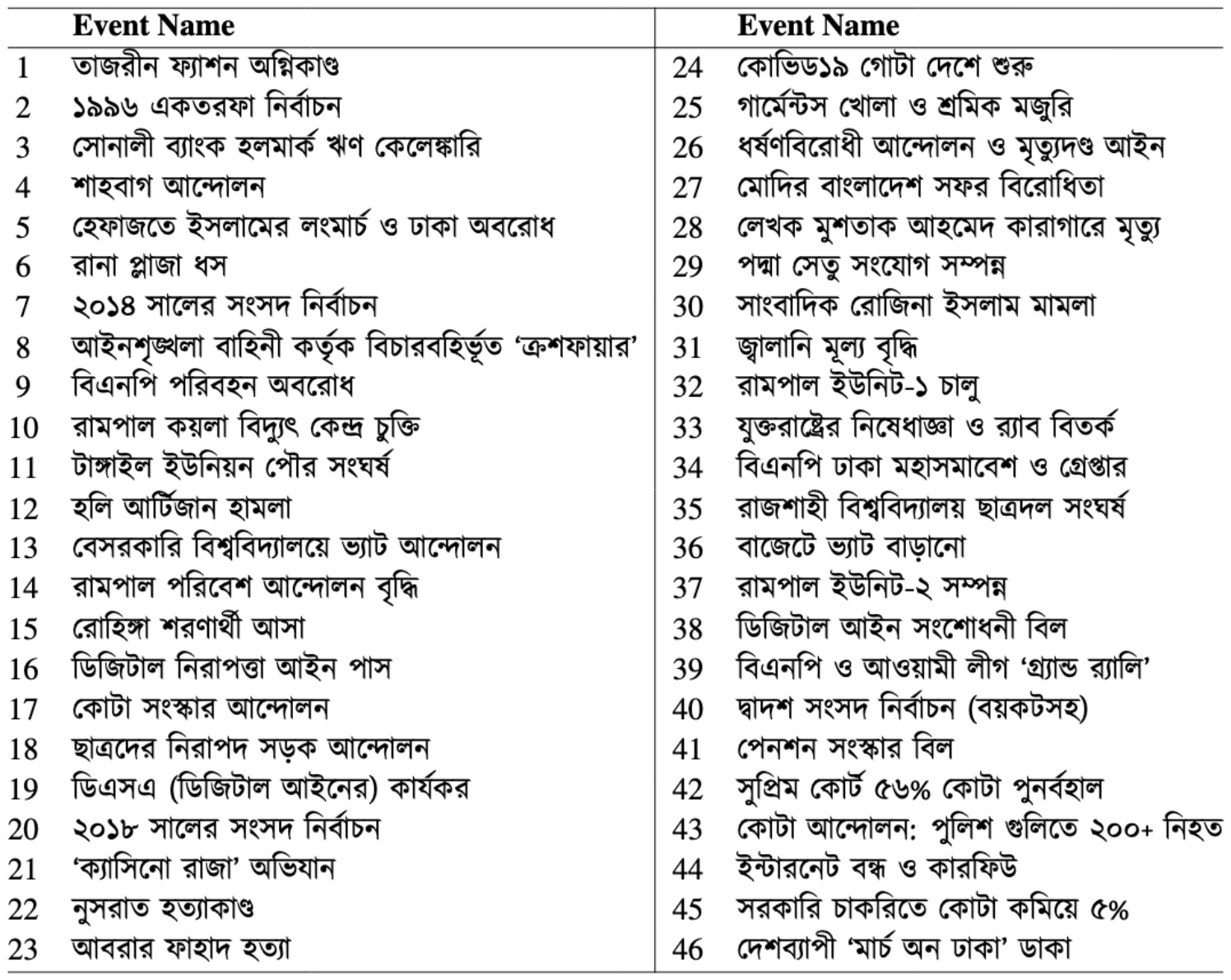} 
    \caption{List of 46 politically debatable events (spanning diverse news coverage) included in our benchmark dataset to capture multiple perspectives.}
    \label{fig:events}
\end{figure*}

\cref{fig:events} includes the list of all the events covered in \ours.
\clearpage

\section{Model ID Mapping}
\cref{tab:model_id_mapping} contains the detailed model id and categories.
\label{sec:model_name_map}
\begin{table*}[ht]
\centering
\begin{tabular}{@{}llr@{}}
\toprule
\textbf{Short Alias} & \textbf{Model ID} & \textbf{Category} \\
\midrule
Qwen2.5-0.5B & Qwen\_Qwen2.5-0.5B-Instruct & Nano \\
Qwen3-0.6B & Qwen\_Qwen3-0.6B & Nano \\
Qwen3-1.7B & Qwen\_Qwen3-1.7B & Nano \\
Qwen2.5-3B & Qwen\_Qwen2.5-3B-Instruct & Compact \\
Qwen3-4B & Qwen\_Qwen3-4B & Compact \\
Qwen3-4B-I & Qwen\_Qwen3-4B-Instruct-2507 & Compact \\
Qwen3-4B-T & Qwen\_Qwen3-4B-Thinking-2507 & Compact \\
Qwen3-8B & Qwen\_Qwen3-8B & Compact \\
Llama3.1-8B & meta-llama\_Llama-3.1-8B-Instruct & Compact \\
TigerLLM-9B & md-nishat-008\_TigerLLM-9B-it & Compact \\
Qwen2.5-14B & Qwen\_Qwen2.5-14B-Instruct & Standard \\
Qwen3-14B & Qwen\_Qwen3-14B & Standard \\
Qwen3-30B-I & Qwen\_Qwen3-30B-A3B-Instruct-2507 & Standard \\
Qwen3-30B-T & Qwen\_Qwen3-30B-A3B-Thinking-2507 & Standard \\
Qwen3-32B & Qwen\_Qwen3-32B & Standard \\
GPT-OSS-20B & openai\_gpt-oss-20b & Standard \\
Llama3.3-70B & meta-llama\_Llama-3.3-70B-Instruct & Massive \\
GPT-OSS-120B & openai\_gpt-oss-120b & Massive \\
Qwen3-235B-I & qwen\_qwen3-235b-a22b-Instruct-2507 & Massive \\
Qwen3-235B-T & qwen\_qwen3-235b-a22b-Thinking-2507 & Massive \\
GLM-4.5 & z-ai\_glm-4.5 & Massive \\
DeepSeek-V3.1 & deepseek\_deepseek-v3.1-terminus & Massive \\
DeepSeek-R1 & deepseek\_deepseek-r1 & Massive \\
Grok-4 & x-ai\_grok-4-fast & Proprietary \\
Claude-Sonnet-4 & anthropic:claude-sonnet-4 & Proprietary \\
Gemini-2.5-Pro & google:gemini-2.5-pro & Proprietary \\
GPT-4.1-Mini & openai:gpt-4.1-mini & Proprietary \\
GPT-5-Mini & openai:gpt-5-mini & Proprietary \\
\bottomrule
\end{tabular}
\caption{Mapping between short aliases and full model id, along with their assigned model category (Nano, Compact, Standard, Massive, or Proprietary).}
\label{tab:model_id_mapping}
\end{table*}

\clearpage

\section{Human Annotator Details}
\label{sec:annotator_info}
Our annotators were graduate students from three different academic backgrounds. Each had a strong understanding of the Bangladesh's socio-political context and actively followed national media coverage. We deliberately selected annotators from diverse majors to foster inclusive knowledge exchange and ensure a multi-perspective approach to the task. None of them had any professional or institutional ties to media outlets. The annotators were fully informed about the purpose of the project and how their annotations would be used.

\section{Evaluation Prompt}
\begin{tcolorbox}[
    coltitle=white,
    colframe=black,
    colback=black!5!white,
    breakable,
    enhanced,
    fontupper=\footnotesize,
    fontlower=\footnotesize,
    fonttitle=\small,
    title={Zero-shot evaluation prompt}
]
\begin{Verbatim}[breaklines=true,breaksymbol=,fontfamily=tt]
You are a political bias classifier.

Task:
Read the following article and return ONLY a valid JSON object with two fields:
{{
  "label": one of ["Govt leaning", "Neutral", "Govt critique"],
  "reason": "1–3 sentences explanation"
}}

Rules:
- Output ONLY JSON.
- Do not include markdown, code fences, or explanations.

Article:
{{article_text}}
\end{Verbatim}
\end{tcolorbox}

\begin{tcolorbox}[
    coltitle=white,
    colframe=black,
    colback=black!5!white,
    breakable,
    enhanced,
    fontupper=\footnotesize,
    fontlower=\footnotesize,
    fonttitle=\small,
    title={Few-shot evaluation prompt}
]
\begin{Verbatim}[breaklines=true,breaksymbol=,fontfamily=tt]
You are a political bias classifier.
 Task:
 Read the following article and return only a valid JSON object with two fields:
{{{{
"label": one of ["Govt leaning", "Neutral", "Govt critique"],
"reason": "1–3 sentences explanation"
}}}}

Rules:
Output ONLY JSON (no markdown, no extra text).


Base your decision strictly on tone, stance, and framing not just factual accuracy.


Examples:
Example 1:
Article:
...
...
Output:
{{"label":"Neutral","reason":"..."}}
Example 2:
Article:
...
...
Output:
{{"label":"Govt leaning","reason":"..."}}
Example 3:
Article:
...
...
Output:
{{"label":"Govt critique","reason":"..."}}

Now classify the following article:
{article_text}
\end{Verbatim}
\end{tcolorbox}
\clearpage
\section{Annotation Decision Flowchart} \label{sec:annotation-dec-flow}
\begin{figure*}[t] 
\includegraphics[width=1\textwidth]{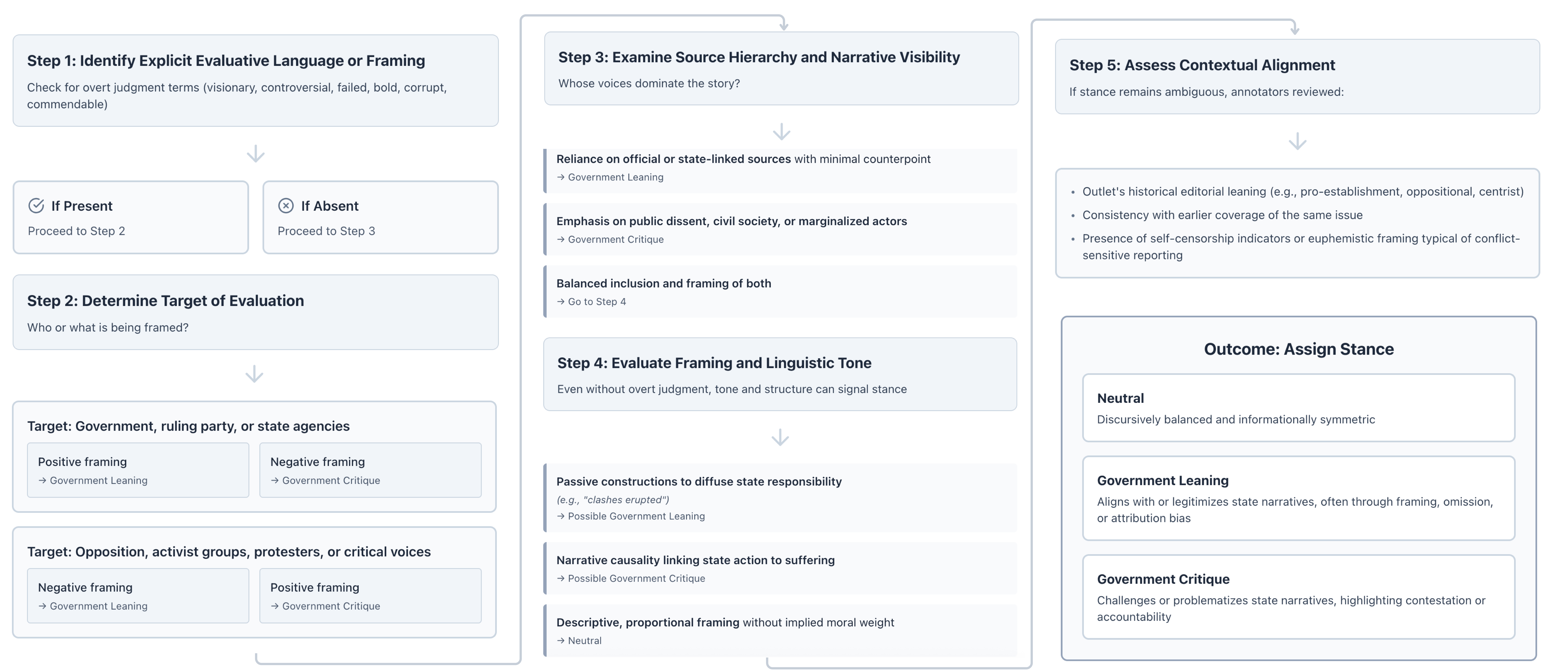} %
\caption{Annotation Decision Flowchart for Political Stance Classification in the Bangladeshi Context. The decision flowchart operationalizes stance annotation within a politically polarized media system, where neutrality is a negotiated discursive position rather than a natural default.}
\label{fig:annotation-flow}
\end{figure*}

A high-level annotation decision flowchart is provided in the \cref{fig:annotation-flow}.

\clearpage
\section{Zero-shot vs Few-shot} \label{sec:prompting-comparison}
Zero-shot vs few-shot prompting is provided in \cref{tab:fewshot_oneshot_weighted_final}.
\begin{table*}[!t]
\centering
\resizebox{\textwidth}{!}{%
\begin{tabular}{@{}l ccc ccc ccc ccc@{}}
\toprule
Model & \multicolumn{3}{c}{Govt. Critique} & \multicolumn{3}{c}{Neutral} & \multicolumn{3}{c}{Govt. Leaning} & \multicolumn{3}{c}{Weighted Avg} \\
\cmidrule(lr){2-4} \cmidrule(lr){5-7} \cmidrule(lr){8-10} \cmidrule(lr){11-13}
 & P & R & F1 & P & R & F1 & P & R & F1 & P & R & F1 \\
\midrule
Qwen3-1.7B (Few-Shot) & 0.66 & 0.78 & 0.71 & 0.56 & 0.60 & 0.58 & 0.55 & 0.18 & 0.27 & 0.60 & 0.62 & 0.59 \\
Qwen3-1.7B (Zero-Shot) & 0.71 & 0.69 & 0.70 & 0.54 & 0.21 & 0.30 & 0.28 & 0.67 & 0.39 & 0.58 & 0.52 & 0.51 \\
\midrule
TigerLLM-9B (Few-Shot) & 0.86 & 0.58 & 0.69 & 0.51 & 0.82 & 0.63 & 0.57 & 0.36 & 0.44 & 0.69 & 0.63 & 0.63 \\
TigerLLM-9B (Zero-Shot) & 0.74 & 0.78 & 0.76 & 0.62 & 0.65 & 0.64 & 0.62 & 0.45 & 0.53 & 0.68 & 0.68 & 0.68 \\
\midrule
Qwen3-32B (Few-Shot) & 0.84 & 0.45 & 0.59 & 0.45 & 0.76 & 0.57 & 0.54 & 0.45 & 0.49 & 0.65 & 0.57 & 0.57 \\
Qwen3-32B (Zero-Shot) & 0.82 & 0.68 & 0.75 & 0.55 & 0.72 & 0.63 & 0.70 & 0.58 & 0.63 & 0.71 & 0.68 & 0.68 \\
\midrule
Llama3.3-70B (Few-Shot) & 0.86 & 0.68 & 0.76 & 0.59 & 0.57 & 0.58 & 0.49 & 0.82 & 0.61 & 0.70 & 0.67 & 0.67 \\
Llama3.3-70B (Zero-Shot) & 0.77 & 0.91 & 0.83 & 0.80 & 0.50 & 0.62 & 0.57 & 0.76 & 0.65 & 0.75 & 0.73 & 0.73 \\
\bottomrule
\end{tabular}%
}
\caption{Few-Shot vs Zero-Shot Performance Comparison}

\label{tab:fewshot_oneshot_weighted_final}
\end{table*}

\end{document}